\documentclass{bmvc2k}
\usepackage{times}
\usepackage{epsfig}
\usepackage{graphicx}
\usepackage{amsmath}
\usepackage{amssymb}
\usepackage{longtable}
\usepackage{multirow}
\usepackage{framed}

\title{PMC-GANs: Generating Multi-Scale High-Quality Pedestrian with Multimodal Cascaded GANs}

\addauthor{Jie Wu}{wujie@cetc-cloud.com}{1}
\addauthor{Ying Peng}{pengying@cetc-cloud.com}{1}
\addauthor{Chenghao Zheng}{zhengchenghao@cetc-cloud.com}{1}
\addauthor{Zongbo Hao}{zbhao@uestc.edu.cn}{2}
\addauthor{Jian Zhang}{zhangjian@cetc-cloud.com}{1}

\addinstitution{
China Electronics Technology Cyber Security Co., Ltd.\\
 Chengdu, China
}
\addinstitution{
School of Information and Software Engineering\\
University of Electronic Science and Technology of China\\
Chengdu, China
}

\runninghead{Wu, Peng, Zheng, Hao, Zhang}{PMC-GANs}

\def\eg{\emph{e.g}\bmvaOneDot}

\begin{document}

\maketitle

\begin{abstract}
Recently, generative adversarial networks (GANs) have shown great advantages in synthesizing images, leading to a boost of explorations of using faked images to augment data. This paper proposes a multimodal cascaded generative adversarial networks (PMC-GANs) to generate realistic and diversified pedestrian images and augment pedestrian detection data. The generator of our model applies a residual U-net structure, with multi-scale residual blocks to encode features, and attention residual blocks to help decode and rebuild pedestrian images. The model constructs in a coarse-to-fine fashion and adopts cascade structure, which is beneficial to produce high-resolution pedestrians. PMC-GANs outperforms baselines, and when used for data augmentation, it improves pedestrian detection results.
\end{abstract}

\section{Introduction}
\label{sec:intro}
\begin{figure}[h]
 \centering
\subfigure{\includegraphics[width=2.4cm]{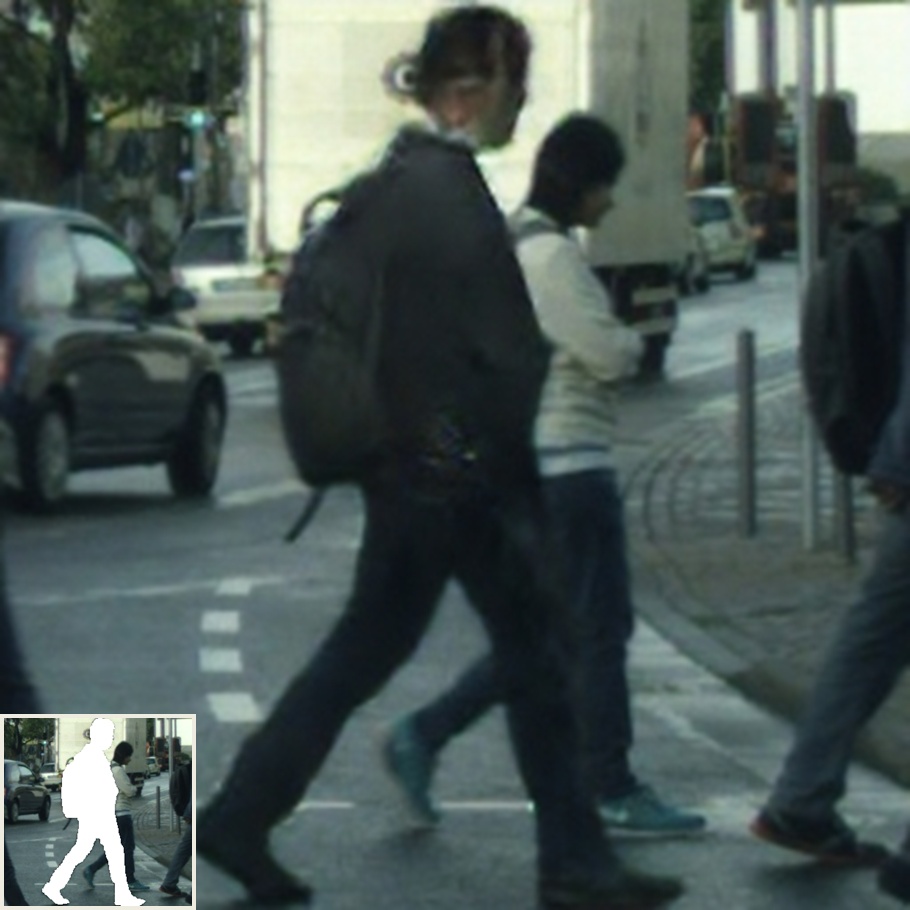}}\,\subfigure{\includegraphics[width=2.4cm]{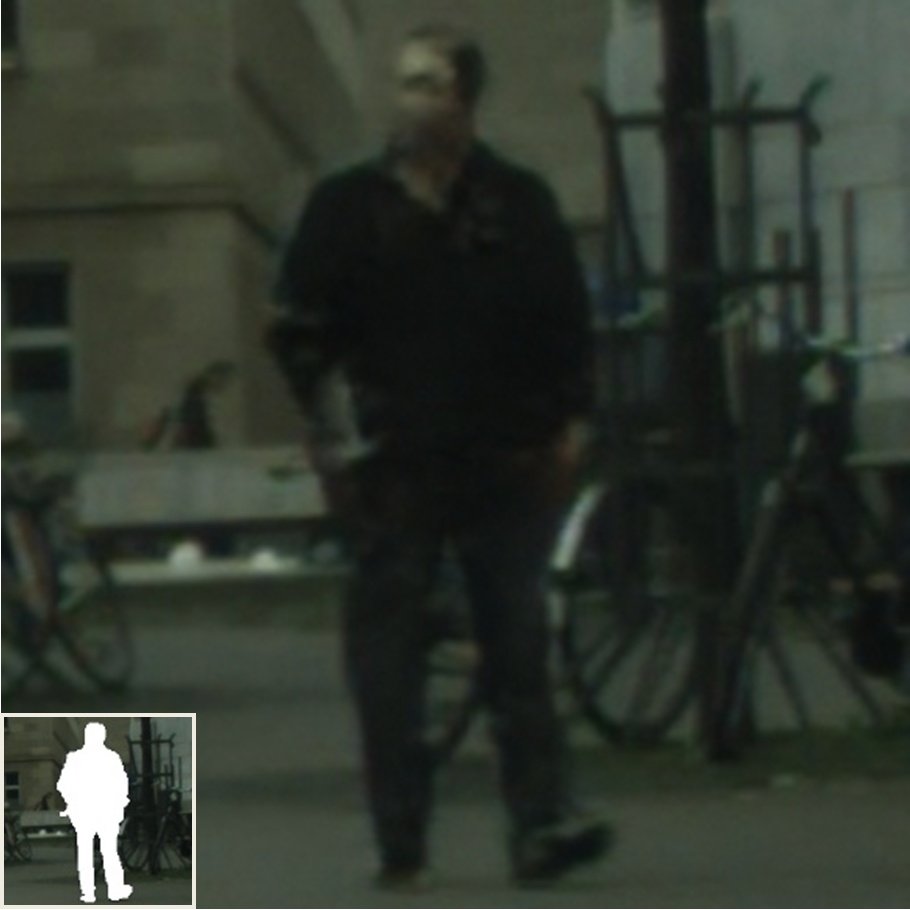}}\,\subfigure{\includegraphics[width=2.4cm]{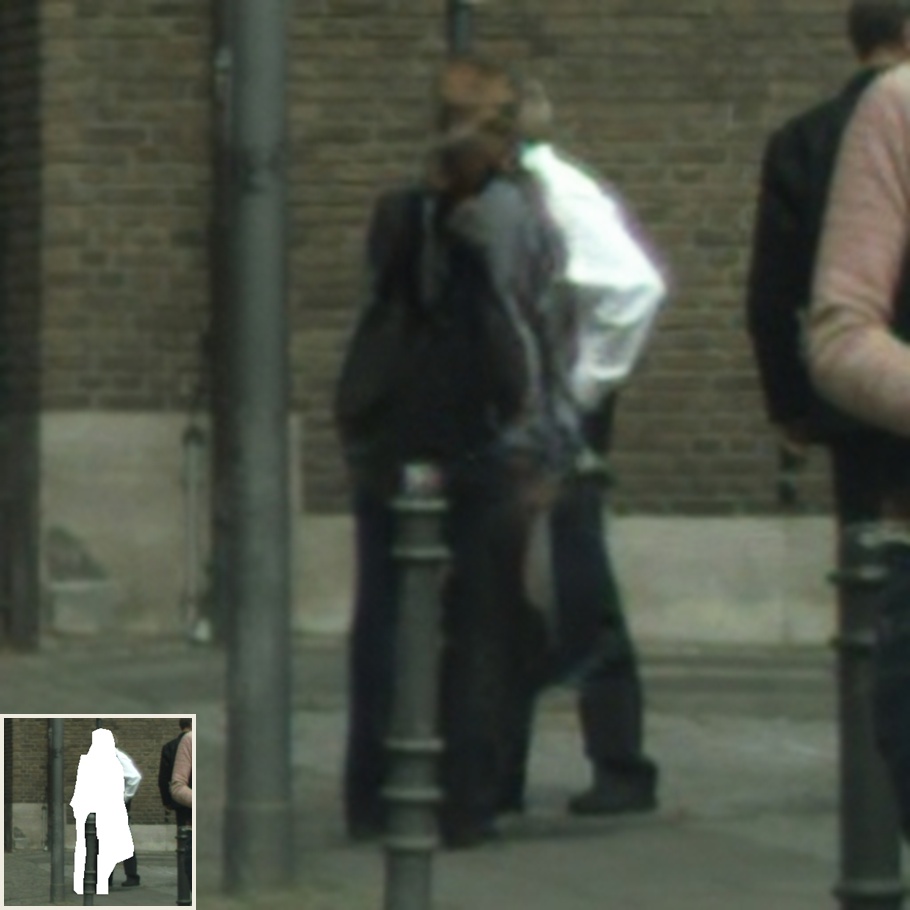}}\,\subfigure{\includegraphics[width=2.4cm]{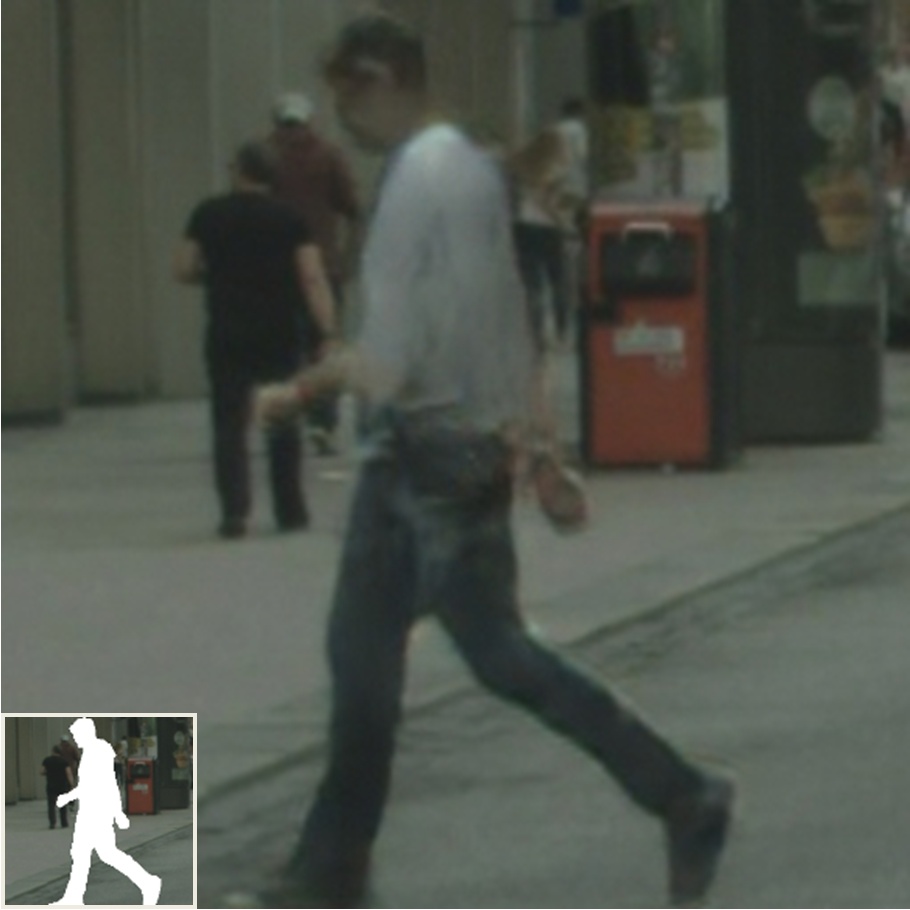}}\,\subfigure{\includegraphics[width=2.4cm]{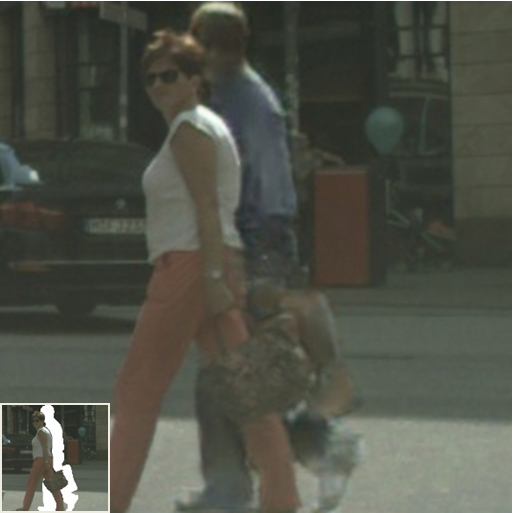}}
 \caption{Pedestrian images generated by our model. The lower left indicates the input.}
 \label{fig 1}
\end{figure}
Pedestrian detection \cite{dollar2014fast},\cite{dollar2009integral},\cite{dollar2009pedestrian},\cite{hosang2015taking},\cite{mao2017can},\cite{zhang2016far} is a fundamental task in applications such as robotics automation, autonomous driving, and video surveillance. In such tasks, training requires a large number of high-quality pedestrian images, but traditional data augmentation methods (\eg, flip and random crop) are not enough, and constructing large-scale manually labeled datasets (\eg, \cite{Cordts2016Cityscapes},\cite{Dollar2012PAMI},\cite{Geiger2012CVPR},\cite{zhang2017citypersons}) are time-consuming and laborious. Therefore, we badly need an efficient way to augment pedestrian data autonomously.

We propose a method based on generative adversarial networks (GANs) to generate high-quality pedestrian images, Figure 1 shows several generated results. GANs \cite{goodfellow2014generative} have recently shown great advantages in synthesizing images \cite{zhu2017unpaired},\cite{zhu2017toward},\cite{wang2017high}. In principle, it is a minimax game, where the generator, $G$, is trained to produce images that are indistinguishable from real ones, and the discriminator, $D$, is trianed to distinguish faked images. The competition between $G$ and $D$ pushes the network to simulate the distribution of real images as closely as possible.

The possibility of using GANs to augment data has been studied in several research areas (\eg, \cite{ZhuAKV18},\cite{frid2018synthetic}, and \cite{mariani2018bagan}), but it is still an open problem in pedestrian detection. Producing visually appealing pedestrians is challenging because of their diverse appearances and sizes. PS-GAN \cite{PSGAN} is the first such attempt, it augments pedestrian detection images with a U-net structured generator \cite{ronneberger2015u}. The network generates a pedestrian within a noise masking rectangular area of the input background image, and adopts a Spatial Pyramid Pooling technique \cite{he2014spatial} to tackle multi-scale pedestrians. However, it has some pitfalls: rectangular masks leave artificial edges when blending with the background; under occlusion, local context information is lost; the details of output pedestrians are rather rough; and the model learns a one-to-one mapping, not efficient enough for data augmentation.

 To produce realistic and diversified high-quality pedestrian images, we make the following improvements: (1) we use instance-level pedestrian masks instead of rectangle masks. Instance-level masks not only indicate the position to generate pedestrians but also inform the shape of pedestrians, thus helpful for generating realistic postures and clear body edges, as well as improves the problem of artificial blending edges and facilitates the synthesizing of occluded pedestrians. (2) We inject a masked latent code into every intermediate block of the encoder part of $G$ to encourage both realistic and diversity of synthesized pedestrians. (3) We upgrade the U-net structured $G$ into a residual U-net, with multi-scale residual blocks in the encoder part and attention residual blocks in the decoder part. Residual blocks \cite{he2016deep} deepen the network and enlarge its capacity. The multi-scale residual blocks further enable modeling multi-scale information. The attention residual blocks help select the most important features in synthesizing images by adjusting the important weight of features. (4) We organize our model into a three-staged cascaded architecture to deal with multi-sized pedestrians. The generators operate at a resolution of $64 \times 64$, $128 \times 128$, and $256 \times 256$, respectively, with a higher stage takes the output of its previous stage as input.

We have experimented on Cityscapes datasets and justified that our PMC-GANs  generates higher-quality pedestrians than baselines. The model is used to augment data for a pedestrian detection task, and the augmented dataset improves the performance of pedestrian detection.

\section{Related Work}
\label{sec 2}
\paragraph{Image-to-Image Translation}
Image-to-image translation learns to map from a source image domain to a target image domain. One of the most famous models is Pix2Pix \cite{isola2017image}, which uses a conditional GAN \cite{mirza2014conditional} with a U-net generator. CycleGAN \cite{zhu2017unpaired} creatively applies cycle consistency losses to improve the reconstruction loss, and the idea of cycle consistency is then widely taken by other studies. These two works learn one-to-one mapping, while our model expects to produce several plausible results with one input. In this field, \cite{almahairi2018augmented} proposes an Augmented CycleGAN, which learns many-to-many mappings by cycling over the original domains augmented with auxiliary latent spaces; \cite{choi2017stargan} designs a star-topology to enable the model to learn multi-domain mappings; \cite{lin2018conditional} produces multi-modal results by regarding the target domain as transforming conditions. We apply a similar fashion as \cite{zhu2017toward}, which learns one-to-many mappings by encouraging a bijection between the output and latent space.
\paragraph{Generate Pedestrians by Using GANs}
\cite{ge2018fd} designs a double-discriminator network to separately distill identity-related and pose-unrelated person features to generate person images. \cite{zhu2019progressive} transfers the pose of a given person to a target pose by applying Pose-Attentional Transfer Blocks within GANs, and \cite{song2019unsupervised} tackles the problem by using adversarial loss together with pose loss, content loss, style loss, and face loss. These works are enlightening, however, their generated pedestrians should maintain the original physical features and costumes of the input image, which is not conducive to the need for diversity in our task. \cite{tripathi2019learning} proposes a composite-based image synthesizing method, which can paste a foreground pedestrian instance into a background image. Compared to this method, our work is able to generate unseen patterns of pedestrians, thus enriching the appearances of augmented pedestrians.
\section{Method}
\label{sec 3}
\begin{figure}
  \centering
  \includegraphics[width=12.7cm]{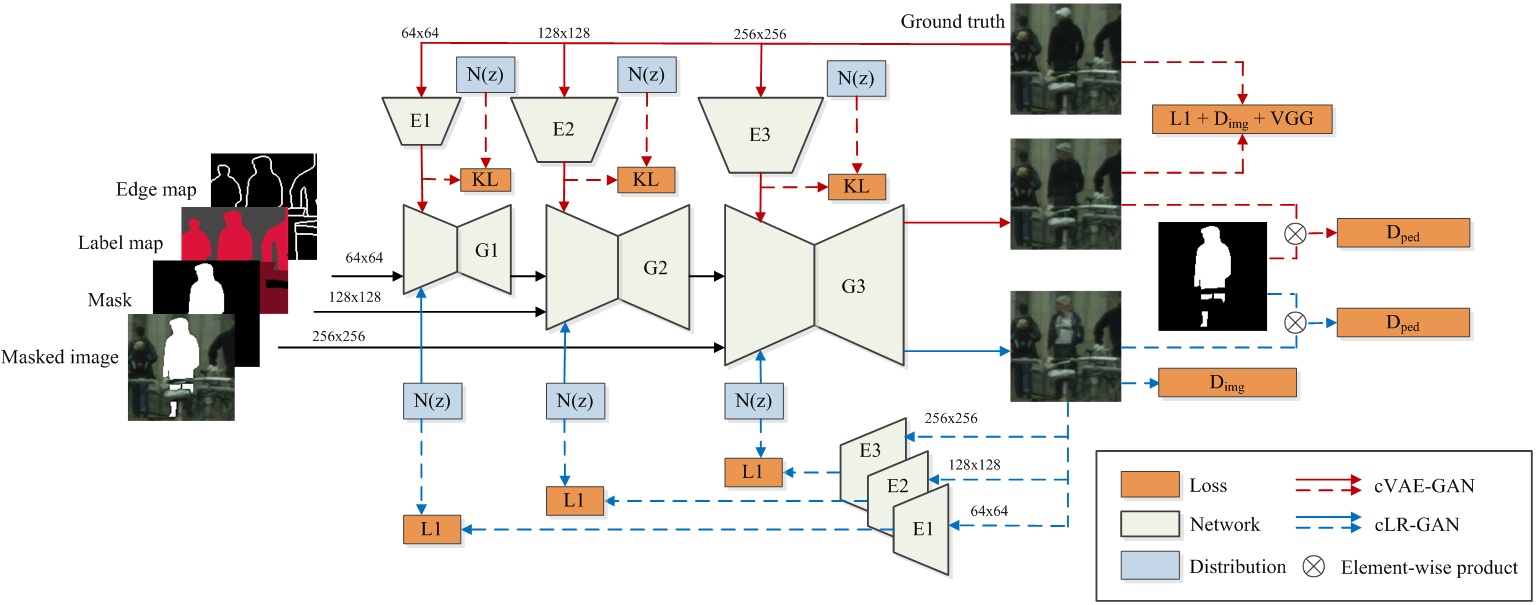}
  \caption{The cascaded architecture of our model. We use the instance-level mask, semantic label map, and edge map (the far left image set) to assist training. Each stage is a hybrid of cVAE-GAN and cLR-GAN, in which cVAE-GAN encodes ground truth by $E_i$, and $G_i$ maps the input along with a sampled $z$ back to ground truth, and cLR-GAN uses a randomly sampled $z$ to map the input into the output domain and then rebuilds $z$ from an output. $D_{img}$ distinguishes real or fake of the whole image, and $D_{ped}$ focuses on the object pedestrian instance.}
  \label{fig 2}
\end{figure}
Our goal is to learn a multi-modal mapping from a source domain, $\mathcal{B}_{M}$, to a target domain, $\mathcal{B}$, where $\mathcal{B}_{M} \subset {\mathbb{R}}^{H \times W \times 3}$ is a masked image domain, and $\mathcal{B} \subset {\mathbb{R}}^{H \times W \times 3}$ is a pedestrian image domain. The pedestrian mask, $\mathcal {M}\subset {\mathbb{R}}^{H \times W \times 1}$, is acquired based on the instance-level semantic label map, $\mathcal {L}_{m}\subset {\mathbb{R}}^{H \times W \times 35}$, of Cityscapes dataset, we set the pixels inside of the objective pedestrian instance to 1 and others to 0. Every $B_M \in \mathcal{B}_{M} $ is computed from a $B \in \mathcal{B}$ and an $M \in \mathcal {M}$  by masking the objective pedestrian instance and leaving the background of $B$ remains. We also use instance-level edge map\cite{wang2017high}, $\mathcal {E}_{m}\subset {\mathbb{R}}^{H \times W \times 1}$, to assist training. During training, we have given a set of paired instances from these domains and the corresponding maps, $\mathcal{A} = \{B_M \in \mathcal{B}_{M}, M \in \mathcal {M}, L_m \in \mathcal {L}_{m}, E_m \in \mathcal {E}_{m} \} $, to represent a joint distribution of $p(A \in \mathcal{A}, B \in \mathcal{B})$. It is important  to note that although there could be multiple plausible $B$ that would fit an input instance $A$, the training set contains only one such pair. Figure 2 is the illustration of the PMC-GANs architecture. During testing, the model is expected to generate a varied set of $\mathcal {B'}$, given an new instance $B_M$. To be specific, the testing instance $B_M$ could be computed either by masking an originally existed pedestrian or by adding a mask at where originally exist no pedestrians.

\subsection{Network Construction}
\label{3.2}
\paragraph{Multi-scale Attention Residual U-net Generator}
Generating high-quality pedestrians is challenging, not only because of the complex body structures but also for the rich and diverse details. Intuitively, we can divide the generation of pedestrians into two steps: encoding, to extract as many infromation as possible from the training instances; decoding, to select the most important features to synthesize pedestrians. Based on this idea, we propose a multi-scale attention residual U-net (U-MAR) structured generator (Figure 3). We introduce multi-scale residual blocks (MSRBs) \cite{li2018multi} into the encoder part of the generator $G$, thus enabling the generator to obtain multi-scale information and to get a robust representation of the input. And we adopt channel attention residual blocks (CARBs) \cite{zhang2018image} to the decoder part of $G$ to adjust the importance weights of features. We find Leaky ReLU activation performs better in our work than ReLU, which is the original settings of these residual blocks. Due to the limit of the length of paper, the inner architecture of MSRBs and CARBs are showed in the supplementary file.
\begin{figure}
  \centering
  \includegraphics[width=12cm]{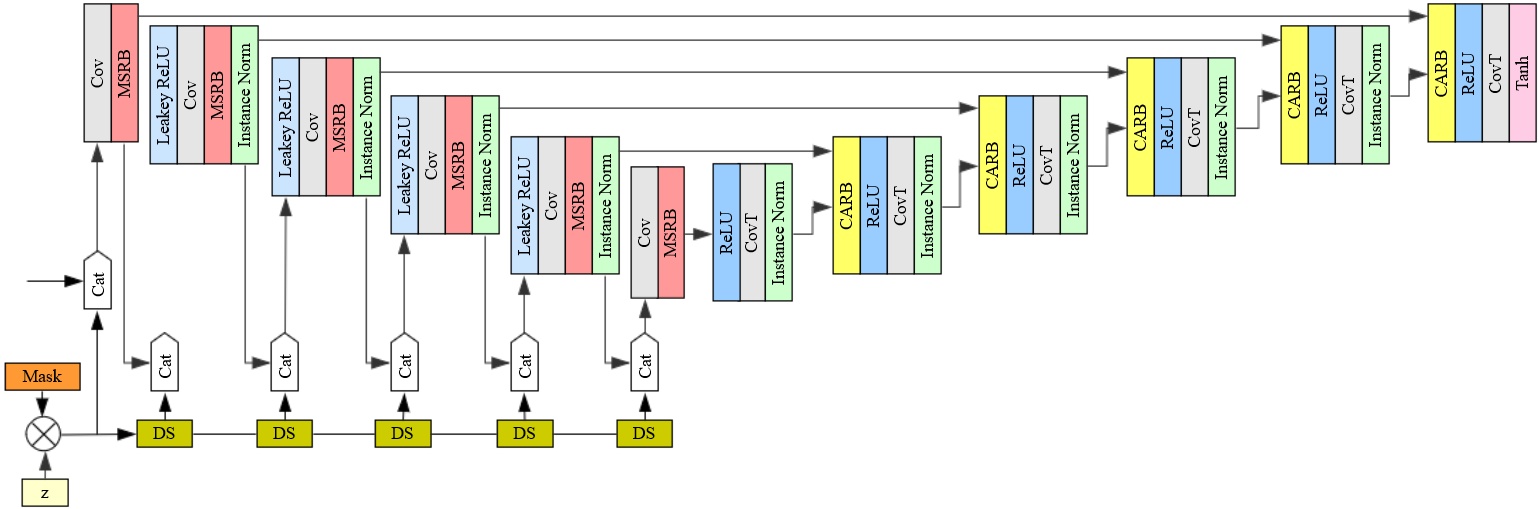}
  \caption{The structure of our generator. $Cov$ is a convolutional layer, $CovT$ is a convolutional transpose layer, $CAT$ is a concatenation function, $DS$ is a downsampling operation, $MSRB$ is a multi-scale residual block, and $CARB$ is a channel attention residual block.}
  \label{fig g-simple}
\end{figure}

\paragraph{Cascaded Architecture}
Generating visually appealing high-resolution pedestrians is much more difficult than generating low-resolution ones, because of finer details. For that differently sized pedestrians share some similar features, such as body structures and textures, it is reasonable to regard the low-resolution result as a start point of training a high-resolution one. We hence organize the network into a cascaded structure, with three stages, operating at resolutions of $64 \times 64$, $128 \times 128$, and $256 \times 256$, respectively. Each stage consists of a GAN that uses the proposed U-MAR structured generators. The generator in a higher stage takes in the integration of the previous stage's knowledge $B'_{previous}$ and the current stage's input $A_{crrent}$. Hence, training a higher stage is also a process of fining its previous. The integration of neighbor stages is showed in Figure 4.
\begin{figure}[t]
  \centering
  \includegraphics[width=11cm]{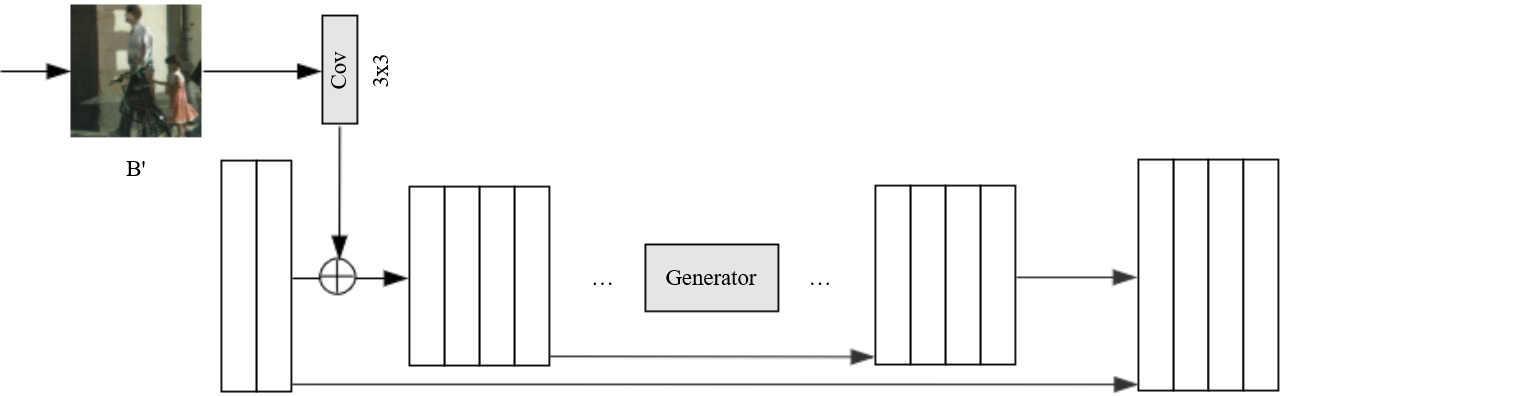}
  \caption{The integration of two stages.}
  \label{fig-cas}
\end{figure}
\subsection{Loss Functions}
\label{3.3}
We train the model based on the settings of BicycleGAN \cite{zhu2017toward}, whose loss function consisted of a conditional LR-GAN loss \cite{donahue2016adversarial},\cite{dumoulin2016adversarially},\cite{chen2016infogan} and a conditional VAE-GAN loss \cite{kingma2014auto}, \cite{larsen2016autoencoding}. The cLR-GAN takes in a random latent code $z$, uses it to map the input $A$ into $B'$, and attempts to rebuild $z$ from $B'$. The cVAE-GAN encodes a ground truth image $B$ into a latent space by an encoder $E$, and the generator tries to map $A$ together with a sampled $z$ back to $B$. Please refer to \cite{zhu2017toward} for more details of BicycleGAN. In our model, the cLR-GAN loss and the cVAE-GAN loss are computed as follows:
\begin{multline}\label{formula 4}
  \mathcal{L}_{GAN} (G,D) = \mathbb{E}_{B'_p \sim p(B'_p)} [log(D(B'_p))]
                               + \mathbb{E}_{A,B'_p\sim p(A,B'_p), z \sim p(z)}[log(1-D(A, G(A, z)))],
\end{multline}
where $\mathcal{L}_{GAN} (G,D)$ is the cLR-GAN loss, $B'_p$ is the product of the generated pedestrian image $B'$ and pedestrian mask $M$. $A$ is the concatenation of an input masked image $B_M $, its corresponding label map $L$, mask $M$, and edge map $E$. $z$ is a random drawn latent code.
\begin{multline}\label{formula 5}
  \mathcal{L}_{GAN}^{VAE} (G,D,E) = \mathbb{E}_{B'_p \sim p(B'_p)} [log(D(B'_p))]
                                + \mathbb{E}_{A,B'_p\sim p(A,B'_p), z \sim E(B'_p)}[log(1-D(A, G(A, z)))],
\end{multline}
where $\mathcal{L}_{GAN}^{VAE} (G,D,E)$ is the cVAE-GAN loss, which modifies equation (1) with a sampling $z \sim E(B'_p)$ by using the re-parameterization trick, allowing for direct back-propagation.

We make two improvements to the loss function: (1) adopt two discriminators, $D_{img}$ and $D_{ped} $, to compute the loss of the whole generated image and the synthesized pedestrian, respectively, while the original BicycleGAN only computes $D_{img}$. (2) Use a perceptual loss \cite{simonyan2014very} based on VGG-19 to encourage the synthesized images to have similar content to the input training instance, i.e., make the output to be more like a pedestrian. The full objective loss function is formulated as:
\begin{multline}\label{formula 6}
   G^{*}, E^{*} =  arg \ \underset{\text{G,E}}{min} \  \underset{\text{D}}{max} {\mathcal{L}}^{VAE}_{GAN} (G,D_{img},E) + \mathcal{L}_{GAN}^{VAE} (G,D_{ped},E) + \lambda{\mathcal{L}}^{VAE}_{1} (G,E)   + {\lambda}_{KL}  {\mathcal{L}}_{KL}(E)  \\
                                + {\mathcal{L}}_{GAN} (G,D_{img}) + \mathcal{L}_{GAN} (G,D_{ped}) + {\lambda}_{latent} {\mathcal{L}}^{latent}_{1} (G,E)
                                + \lambda_{VGG} {\mathcal{L}}_{VGG}(G,E),
\end{multline}
where ${\mathcal{L}}^{VAE}_{GAN} (\cdot)$ and ${\mathcal{L}}_{GAN} (\cdot)$ are adversarial losses of cVAE-GAN and cLR-GAN, respectively.  ${\mathcal{L}}^{VAE}_{1} (\cdot)$ is $\mathcal{L}1$ loss between image $B$ and $B'$, driving $G$ to match $B$. ${\mathcal{L}}^{latent}_{1} (\cdot)$ encourages $E$ to produce a latent code that is close to a Gaussian distribution.  ${\mathcal{L}}_{KL}(E)$ is $KL$ distance in cLR-GAN. The hyper-parameters $ \lambda$, $\lambda_{KL}$,  ${\lambda}_{latent}$, and ${\lambda}_{VGG}$ control the relative importance of each term.
\section{Experiments}
\label{sec 4}
\paragraph{Implementation Details}
At each stage, $G$ uses the proposed U-MAR structure, $D_{img}$ and $D_{ped}$ use the PatchGAN \cite{isola2017image}, and $E$ uses the ResNet\cite{he2016deep}. LSGANs \cite{mao2017least} is adopted to compute adversarial losses. A 16-dimensional latent code $z$ is injected into the network by spatial replication, multiplied with the pedestrian mask, down-sampled by the nearest neighbor interpolation, and then concatenation into every intermediate layer of the encoder part of the generator. The parameters, $\lambda$, ${\lambda}_{latent}$, ${\lambda}_{KL}$ and ${\lambda}_{VGG}$ are set to 10, 0.5, 0.01, and 1, respectively. We set the batch size to 1, and the epoch to 200. The perceptual loss is \emph{not} used in the first stage for it results in unstable during training. When training a cascaded model, the first stage is trained in a similar setting as in \cite{zhu2017toward} to learn $G_1$ and $E_1$. The second stage trains the first 100 epochs for $G_2$ and $E_2$ with fixed $G_1$ and $E_1$, and trains another 100 epochs on all of the $G_1$, $G_2$, $E_1$, and $E_2$. The third stage uses the same strategy as the second one, and trains on $G_2$, $G_3$, $E_2$, and $E_3$. We use Adam \cite{kinga2015method} optimizer, with a learning rate of $w^{h-i}* lr$, where $lr$ is the basic learning rate, $h$ is the total number of cascaded stages, $i$ is the ordinal of the current training stage, and $w$ is a weight factor. We set $lr$ to 0.0002 and $w$ to 0.01. By fixing $G$ and $E$ of a previous stage as well as setting a small factor of $w$, we force a higher stage to better follow the already learned knowledge of its previous ones.

\paragraph{Dataset}
The model is trained on the Cityscapes training set and evaluated on its validation set. The size $s_i$ of input images in the $i^{th}$ stage are set to $s_1 = 64$, $s_2 = 128$, and $s_3 = 256$. We crop the pedestrian images from Cityscapes dataset, every image shares the same center with the corresponding pedestrian bounding box. Let $H$ denotes the height of the bounding box of a pedestrian. At the $i^{th}$ stage, if a $H$ is smaller than $s_i$, the corresponding pedestrian image is cropped at the resolution of $s_i \times s_i$, and if $H$ is bigger than $s_i$, the image is cropped at a resolution of $H \times H$, and then resized to $s_i \times s_i$. Considering that resizing an image too much can lead to information loss, we limit $H$ of the first staged pedestrians to 64 to 256, the second to 100 to 1024, and the third to 150 to 1024. Because high-resolution pedestrians are fewer than low-resolution ones, the last stage contains far fewer images than the other two. Therefore, we expand the training set of the third stage as follows: for every pedestrian image, randomly pick a value from the interval, (H, 1.22 H], to be $H'$, and then crop on the resolution of $H' \times H'$ and resize to $s_i \times s_i$. The training set of each stage contains 6,000 images, 4,700 images, and 5,600 images, respectively, and the validation sets contain 1,000 images each.

\paragraph{Evaluation Metrics} We use the Fr{\'e}chet Inception Distance (FID) measurement \cite{heusel2017gans}, which calculates the distance between generated images and ground truth images in the Inception-v3 network feature space \cite{zhang2018self}. The score of FID is consistent with human judgment \cite{heusel2017gans}, which rewards realistic synthesized images and penalizes a lack of diversity \cite{brock2018large}. The formula follows \cite{dowson1982frechet},\cite{heusel2017gans}, with the lower FID score, the better. For every input, we generate a set of samples by using 5 random latent codes, and take them as a whole to compute FID score.

\paragraph{Baselines}
Two previous works: (1) \cite{wang2017high}, a pix2pix-based model, uses generators with residual blocks, and takes instance-level masks and edge maps as the aid of input. We only use its \emph{global generator} in the experiments. (2) \cite{PSGAN}, a pedestrian synthesis model, based on pix2pix network and uses U-net generator. And four ablation versions of our work: (1) Ours-1, uses basic U-net generator. (2) Ours-2, uses basic residual blocks to replace the multi-scale and the attention ones in the proposed generator. (3) Ours-3, uses MSRBs in the encoder part of the generator, and uses basic residual blocks in the decoder part. (4) Ours-4, uses the proposed generator structure. For fair, all the baseline models are trained for 200 epochs, optimized by Adam, and reduce the learning rate from the $100^{th}$ epoch. The baseline methods are trained in a \emph{one-staged-fashion} without cascaded architecture. Each method is trained on three input resolutions, based on our dataset. \cite{wang2017high} and \cite{PSGAN} use their original loss functions, and the other baselines use the same loss function as our PMC-GANs.

\subsection{Results}
\paragraph{Qualitative Comparison} Figure 5 shows the output images of baseline methods and our work, at the resolutions of $64 \times 64$, $128 \times 128$, and $256 \times 256$ in the three rows. The model, Ours (PMC-GANs) generates a more delicate pedestrian at every one of the resolutions, the details of the pedestrians are richer, and the body part boundaries are clearer. Our model learns a multimodal mapping, which improves the efficiency of data augmentation as well as avoids mode collapse. Figure 6 shows some multimodal results.
\begin{figure}[h]
  \centering
\subfigure{\includegraphics[width=11.7cm]{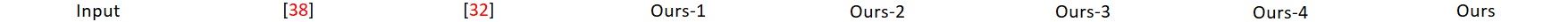}}\vspace{-0.3cm}
\subfigure{\includegraphics[width=1.4cm]{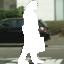}}\,\subfigure{\includegraphics[width=1.4cm]{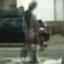}}\,\subfigure{\includegraphics[width=1.4cm]{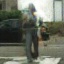}}\,\subfigure{\includegraphics[width=1.4cm]{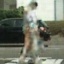}}\,\subfigure{\includegraphics[width=1.4cm]{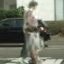}}\,\subfigure{\includegraphics[width=1.4cm]{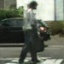}}\,\subfigure{\includegraphics[width=1.4cm]{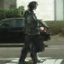}}\,\subfigure{\includegraphics[width=1.4cm]{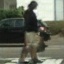}}\vspace{-0.3cm}
\subfigure{\includegraphics[width=1.4cm]{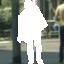}}\,\subfigure{\includegraphics[width=1.4cm]{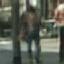}}\,\subfigure{\includegraphics[width=1.4cm]{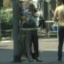}}\,\subfigure{\includegraphics[width=1.4cm]{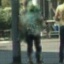}}\,\subfigure{\includegraphics[width=1.4cm]{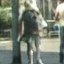}}\,\subfigure{\includegraphics[width=1.4cm]{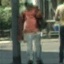}}\,\subfigure{\includegraphics[width=1.4cm]{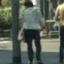}}\,\subfigure{\includegraphics[width=1.4cm]{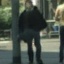}}\vspace{-0.3cm}
\subfigure{\includegraphics[width=1.4cm]{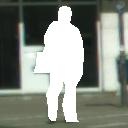}}\,\subfigure{\includegraphics[width=1.4cm]{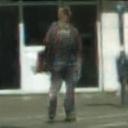}}\,\subfigure{\includegraphics[width=1.4cm]{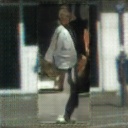}}\,\subfigure{\includegraphics[width=1.4cm]{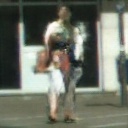}}\,\subfigure{\includegraphics[width=1.4cm]{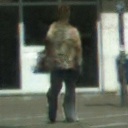}}\,\subfigure{\includegraphics[width=1.4cm]{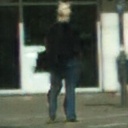}}\,\subfigure{\includegraphics[width=1.4cm]{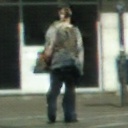}}\,\subfigure{\includegraphics[width=1.4cm]{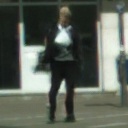}}\vspace{-0.3cm}
\subfigure{\includegraphics[width=1.4cm]{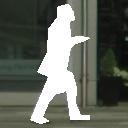}}\,\subfigure{\includegraphics[width=1.4cm]{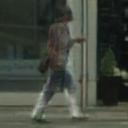}}\,\subfigure{\includegraphics[width=1.4cm]{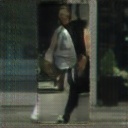}}\,\subfigure{\includegraphics[width=1.4cm]{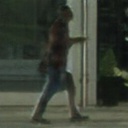}}\,\subfigure{\includegraphics[width=1.4cm]{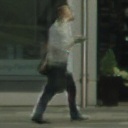}}\,\subfigure{\includegraphics[width=1.4cm]{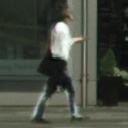}}\,\subfigure{\includegraphics[width=1.4cm]{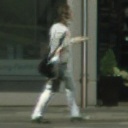}}\,\subfigure{\includegraphics[width=1.4cm]{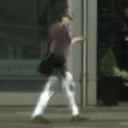}}\vspace{-0.3cm}
\subfigure{\includegraphics[width=1.4cm]{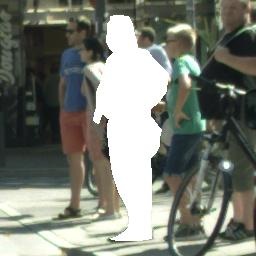}}\,\subfigure{\includegraphics[width=1.4cm]{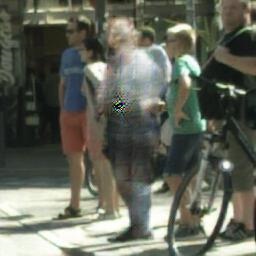}}\,\subfigure{\includegraphics[width=1.4cm]{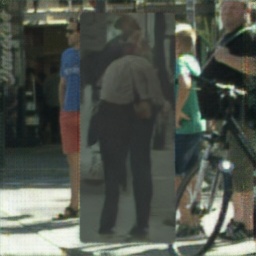}}\,\subfigure{\includegraphics[width=1.4cm]{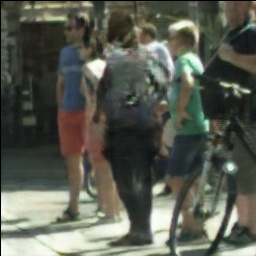}}\,\subfigure{\includegraphics[width=1.4cm]{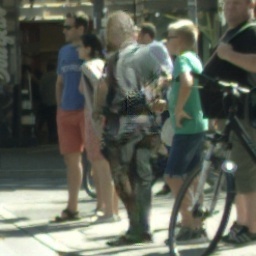}}\,\subfigure{\includegraphics[width=1.4cm]{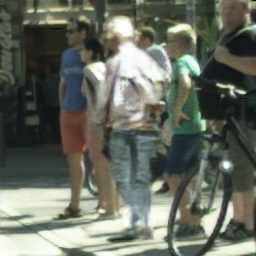}}\,\subfigure{\includegraphics[width=1.4cm]{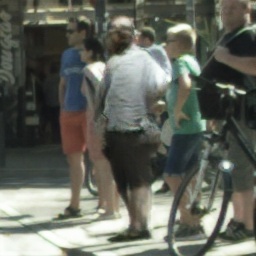}}\,\subfigure{\includegraphics[width=1.4cm]{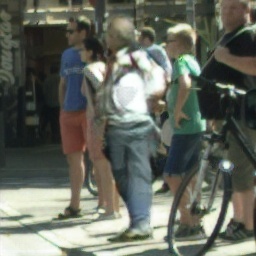}}\vspace{-0.3cm}
\subfigure{\includegraphics[width=1.4cm]{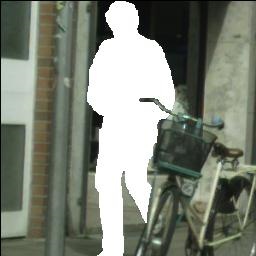}}\,\subfigure{\includegraphics[width=1.4cm]{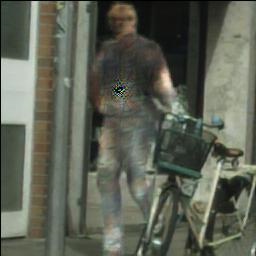}}\,\subfigure{\includegraphics[width=1.4cm]{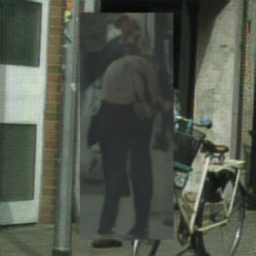}}\,\subfigure{\includegraphics[width=1.4cm]{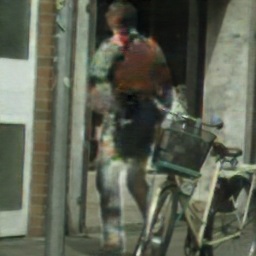}}\,\subfigure{\includegraphics[width=1.4cm]{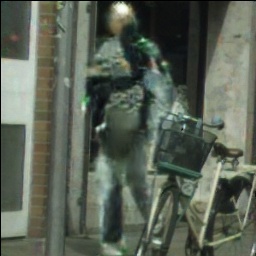}}\,\subfigure{\includegraphics[width=1.4cm]{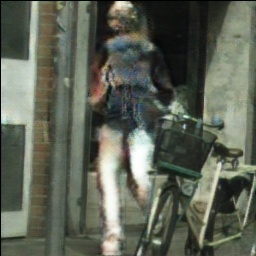}}\,\subfigure{\includegraphics[width=1.4cm]{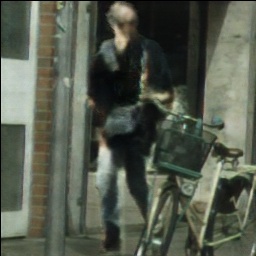}}\,\subfigure{\includegraphics[width=1.4cm]{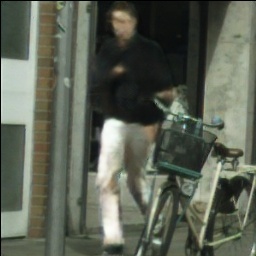}}
\caption{Visual comparison of PMC-GANs with baselines. The first two lines are generated at a resolution of $64 \times 64$, the third and fourth lines are results at a resolution of $128 \times 128$, and the last two lines are generated at a resolution of $256\times 256$.}
 \label{fig 5}
\end{figure}
\begin{figure}[h]
  \centering
\subfigure{\includegraphics[width=1.22cm]{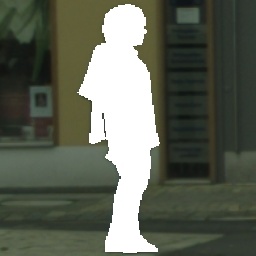}}\,\subfigure{\includegraphics[width=1.22cm]{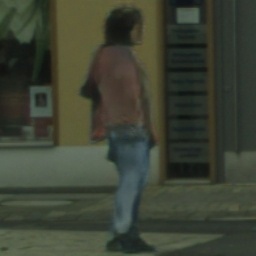}}\,\subfigure{\includegraphics[width=1.22cm]{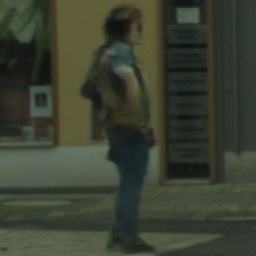}}\,\subfigure{\includegraphics[width=1.22cm]{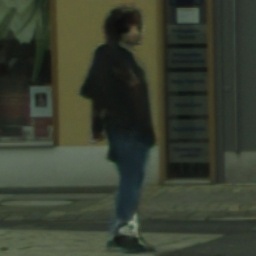}}\,\subfigure{\includegraphics[width=1.22cm]{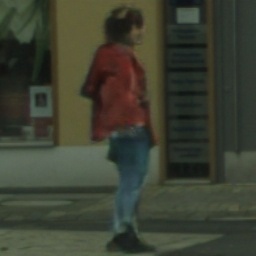}}\,\subfigure{\includegraphics[width=1.22cm]{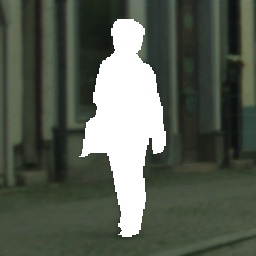}}\,\subfigure{\includegraphics[width=1.22cm]{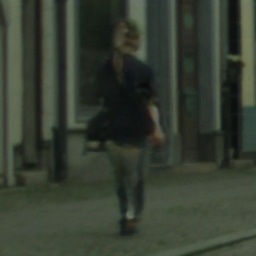}}\,\subfigure{\includegraphics[width=1.22cm]{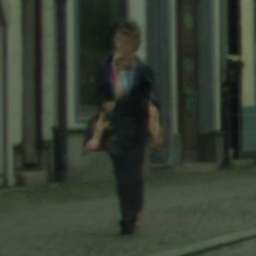}}\,\subfigure{\includegraphics[width=1.22cm]{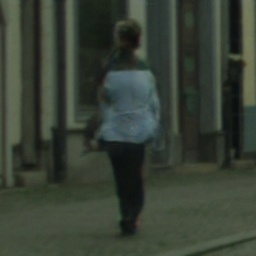}}\,\subfigure{\includegraphics[width=1.22cm]{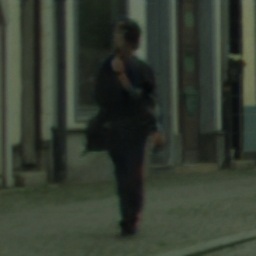}}\vspace{-0.3cm}
\subfigure{\includegraphics[width=1.22cm]{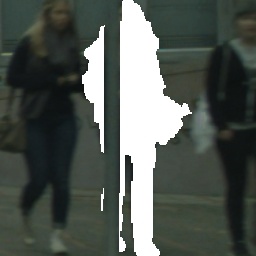}}\,\subfigure{\includegraphics[width=1.22cm]{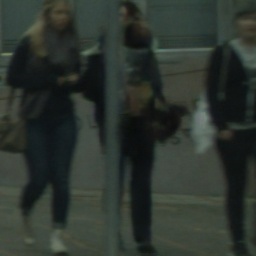}}\,\subfigure{\includegraphics[width=1.22cm]{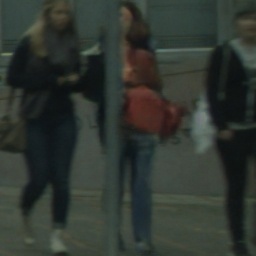}}\,\subfigure{\includegraphics[width=1.22cm]{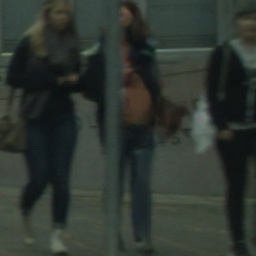}}\,\subfigure{\includegraphics[width=1.22cm]{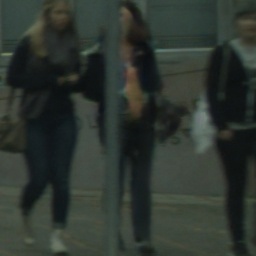}}\,\subfigure{\includegraphics[width=1.22cm]{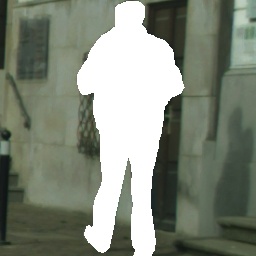}}\,\subfigure{\includegraphics[width=1.22cm]{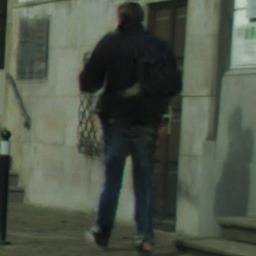}}\,\subfigure{\includegraphics[width=1.22cm]{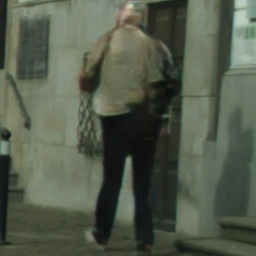}}\,\subfigure{\includegraphics[width=1.22cm]{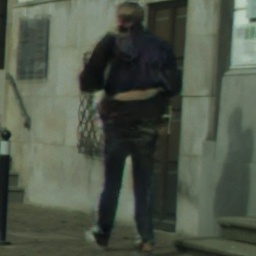}}\,\subfigure{\includegraphics[width=1.22cm]{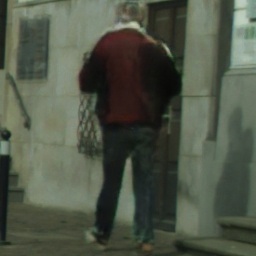}}
\caption{The multimodal products of PMC-GANs.}
\label{fig mul}
\end{figure}

We perform an interpolation experiment on PMC-GANs by manipulating the random code $z$, which is injected into the generator. Figure 7 shows an interpolation instance, where the first and the last generated images are produced by randomly sampling a $z_{first}$ and a $z_{last}$ from a Gaussian distribution, and the others are produced by injecting the interpolate values between $z_{first}$ and $z_{last}$ to the generator. The model produces different images on neighbor interpolation samples, illustrating that it does not over-fit to the training data.
\begin{figure}[t]
  \centering
\subfigure{\includegraphics[width=1.22cm]{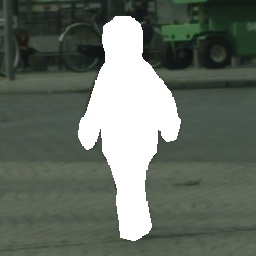}}\,\subfigure{\includegraphics[width=1.22cm]{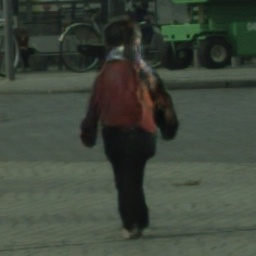}}\,\subfigure{\includegraphics[width=1.22cm]{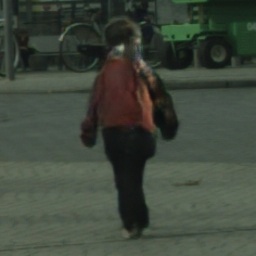}}\,\subfigure{\includegraphics[width=1.22cm]{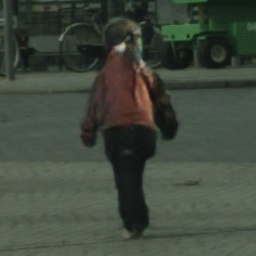}}\,\subfigure{\includegraphics[width=1.22cm]{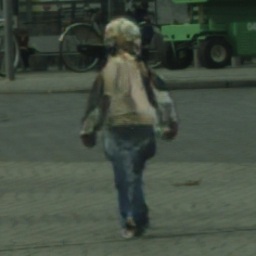}}\,\subfigure{\includegraphics[width=1.22cm]{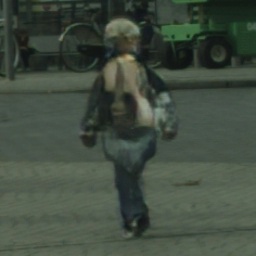}}\,\subfigure{\includegraphics[width=1.22cm]{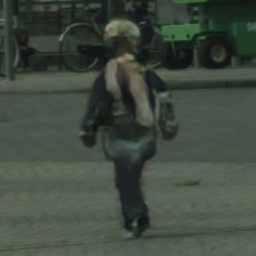}}\,\subfigure{\includegraphics[width=1.22cm]{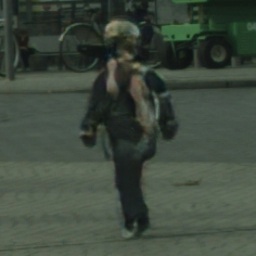}}\,\subfigure{\includegraphics[width=1.22cm]{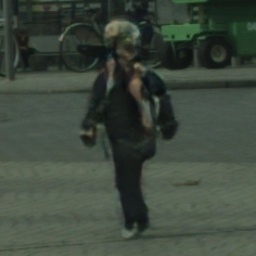}}\,\subfigure{\includegraphics[width=1.22cm]{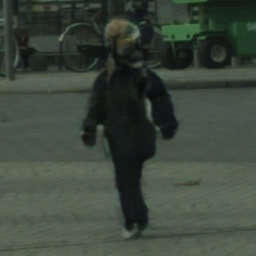}}
  \caption{An interpolation analysis.}
  \label{fig 7}
\end{figure}
\paragraph{Quantitative Evaluation}
We upgrade a U-net generator into a U-MAR generator, so the model, Ours-1, is a basic baseline of our study. We perform an ablation study based on Ours-1 to justify the effectiveness of $L_m$, $E_m$, $D_{ped}$, and $\mathcal{L}_{VGG}$ in synthesizing pedestrians. The experiment is conducted at a resolution of $256 \times 256$, all the models are trained in a one-staged-fashion by using the same settings as Ours-1. $L_m$ and $E_m$ are always bundled to each other, so we use $LE_m$ to denote the co-occurrence of the two. Table 1 shows that cutting out any of the variables will lead to a worse FID score. Therefore, it is reasonable to use $L_m$, $E_m$, $D_{ped}$, and $\mathcal{L}_{VGG}$ in our work.
\begin{table}[h]
\centering
\caption{The results of the ablation study. The best performance is in \textbf{bold}.}
\label{tab abl}
\begin{tabular}{l|ccc|c}
\hline
Model &  $LE_m$ & $D_{ped}$ & $\mathcal{L}_{VGG}$ & FID\\
\hline
Ours-$1^{*}$     & $\times$ &$\checkmark$ &$\checkmark$ &   30.74\\
Ours-$1^{**}$  & $\checkmark$ & $\times$ & $\checkmark$&   38.24\\
 Ours-$1^{***}$ & $\checkmark$ & $\checkmark$ & $\times$&  36.66\\
Ours-1   & $\checkmark$ & $\checkmark$ & $\checkmark$ & \textbf{30.12} \\
\hline
\end{tabular}
\end{table}

Table 2 shows FID scores of the baseline methods and our model. By comparing \cite{PSGAN} and \cite{wang2017high} with the ablation versions of our model, we can see our work gets a better result on the whole, which is consistent with qualitative comparison results. Ours-4, with the proposed generator, improves the FID score by  7.6\%, 27.2\%, and 17.7\%, on Ours-1, at the resolution of $64\times 64$, $128\times128$, and $256\times 256$, respectively, indicating the superior of the U-MAR generator than the basic U-net generator in producing pedestrians. The comparison between Ours-1 and Ours-2 shows the advance of adding residual blocks in basic U-net structure in our task, the comparison between Ours-2 and Ours-3 validates the usefulness of using the MSRBs in the encoder part of the generator, and the comparison between Ours-3 and Ours-4 justifies the effectiveness of using the CARBs in the decoder part of the generator.

The model, Ours, and the model, Ours-4, use the same structured $G$, $D$, and $E$, the difference between the two is that Ours uses cascaded architecture to produce high-resolution images, while Ours-4 uses only one stage. At the resolution of $64 \times 64$, Ours ablates to Ours-4, and the higher the resolution, the greater the advantage of the cascaded structure is revealed, with a benefit of improving 0.7\% at the resolution of $128 \times 128$, and 9.3\% at $256 \times 256$. The results show that a coarse-to-fine cascaded architecture is useful in synthesizing high-resolution pedestrian images.
\begin{table}[h]
  \centering
  \caption{Comparison of FID score. The best results are in \textbf{bold}.}
  \label{tab 1}
  \begin{tabular}{c|cc|cccc|c}
  \hline
  \multirow{2}{1.5cm}{Resolution} & \multicolumn{6}{c}{Models}\\
  \cline{2-8}
   & \cite{wang2017high}&  \cite{PSGAN} &Ours-1 & Ours-2 &Ours-3&Ours-4 &Ours\\
   \hline
  $64\times 64$ & 20.46  & 27.92 & 10.91 &10.76 & \textbf{10.03} & 10.08  & 10.08 \\
  \hline
  $128\times 128$ & 27.69 & 78.67 & 23.11 &18.69 & 17.30 & 16.82 & \textbf{16.71}\\
  \hline
 $256 \times 256$ & 39.13 & 82.14 &  30.12 & 27.66 & 34.37 & 24.79 &\textbf{22.48} \\
  \hline
  \end{tabular}
  \end{table}

\subsection{Data Augmentation Experiments}
\begin{figure}[h]
  \centering
\subfigure{\includegraphics[width=6.3cm]{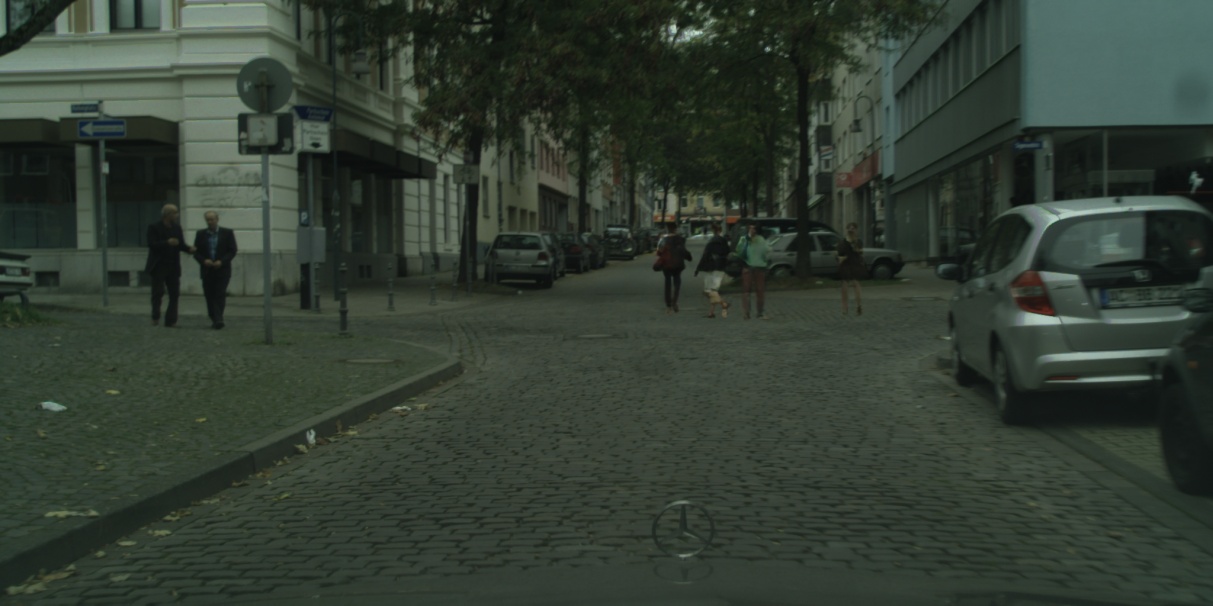}}\,\subfigure{\includegraphics[width=6.3cm]{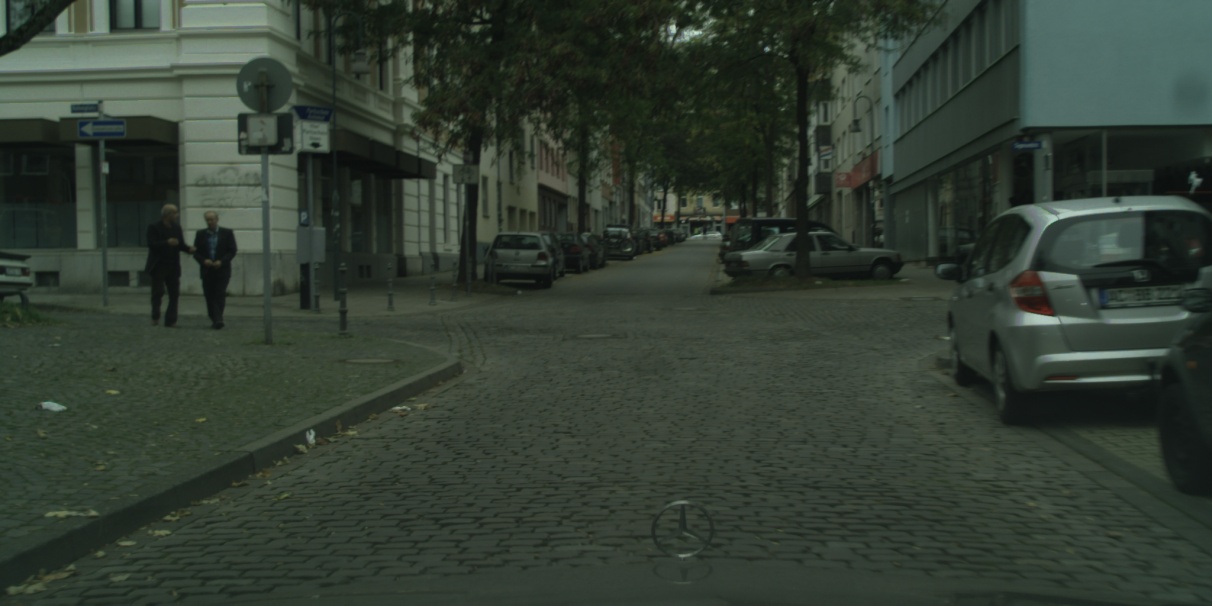}}\vspace{-0.3cm}
\subfigure{\includegraphics[width=6.3cm]{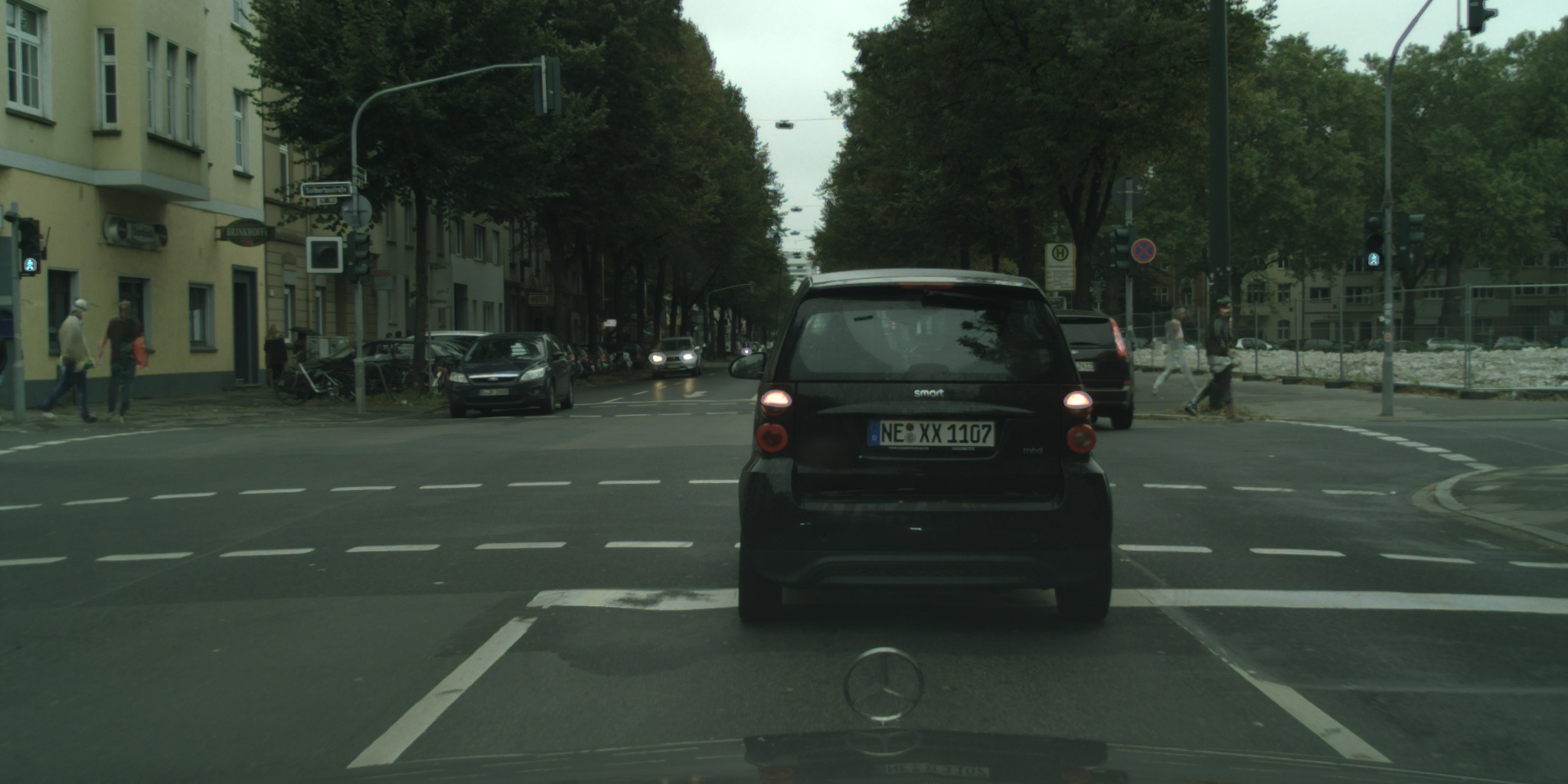}}\,\subfigure{\includegraphics[width=6.3cm]{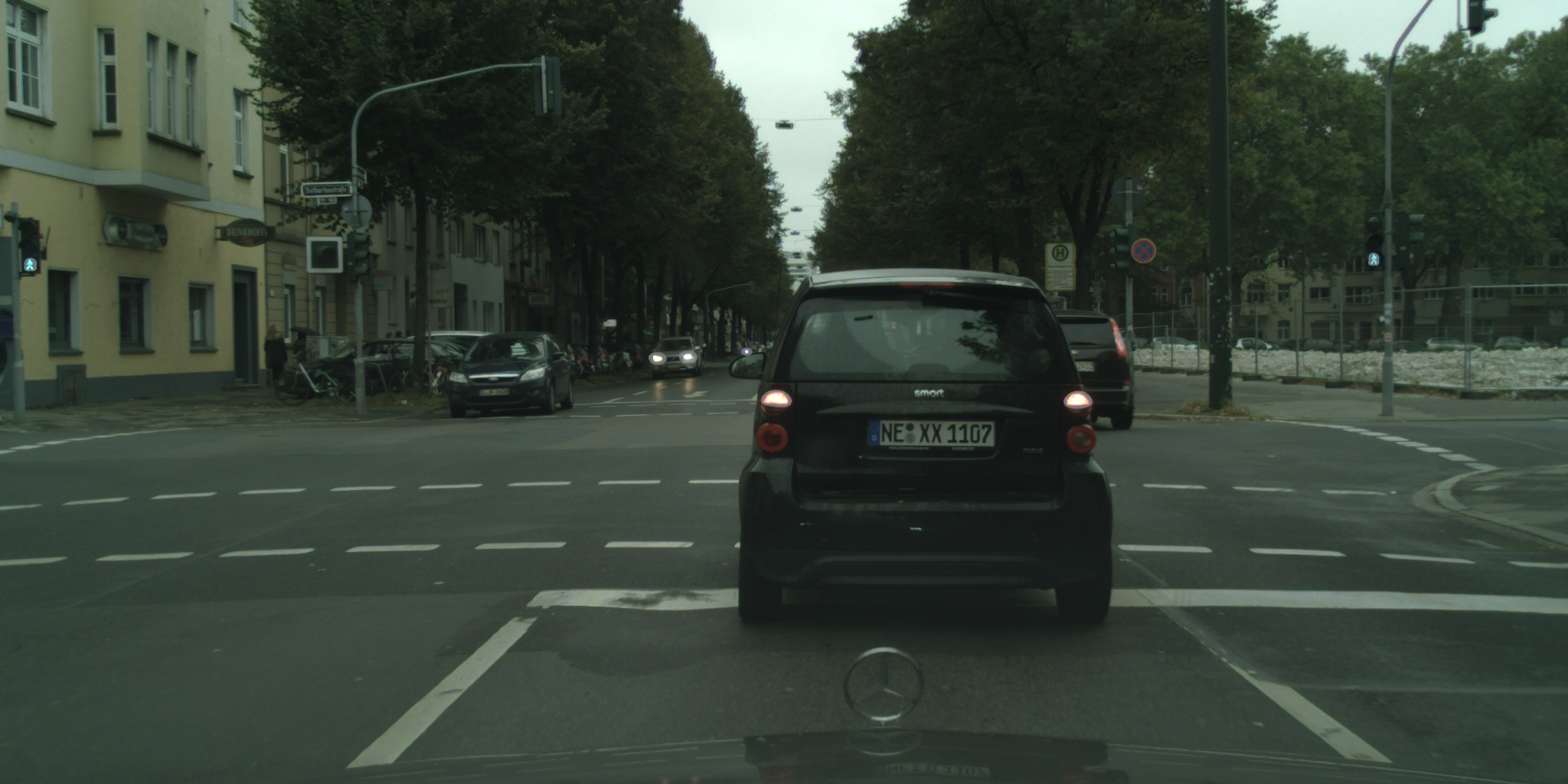}}
  \caption{Data augmentation samples, where the left ones are synthesized images.}
  \label{fig 8}
\end{figure}
Experiments are performed on the CityPersons dataset \cite{zhang2017citypersons}. We decide the position $P_{p}$ and the size $P_{s}$ of a synthesized pedestrian by first, use semantic label map to restrict the position of it to sidewalks and roads; then, compute the size of pedestrians according to both the size of existing cars and pedestrians in the image and the distribution of pedestrian size conditioned on the position in the dataset. We crop a $P_{s} \times P_{s}$ sized background image, $I_{bg}$, centered at $P_{p}$, from an CityPersons image, $I_{HD}$. Then randomly select a pedestrian mask $M$ and compute with $I_{bg}$ to acquire the masked image input to our trained PMC-GANs. The generated pedestrian image $I_{ped}$ is then blende into $I_{HD}$ by using a pix-wise replacement strategy. We then select 3,000 blended images, the same to the quantity of the CityPersons training set, as the augmented data (Figure 8 shows two augmented samples). The tested pedestrian detector uses ResNet-50\cite{he2016deep} as the backbone network and pre-trained on the ImageNet dataset\cite{russakovsky2015imagenet}. Pedestrians with a height less than 50 pixels or the visible ratio less than 0.65 are \emph{ignored}. During training both the RPN proposals and the Fast R-CNN\cite{girshick2015fast} stage, we avoid sampling the ignore regions. Random crop and horizontal flip strategies are also applied to augment data. We use $\times 1$ images to train the $\times 1$ detector, and use the scale of $\times 1.3$ upsampling images to train the $\times 1.3$ detector. To fit in 12GB of TITAN X GPU memory and avoid memory overflow, the $\times 1.5$ scaled detector is trained on $\times 1.3$ scaled images and tested on the $\times 1.5$ dataset. Please see supplementary file for more information. Table 3 shows the comparison of our pedestrian detection method with baselines on the CityPersons validation set, which justifies that augmenting data by using PMC-GANs is effective at every testing scales.
\begin{table}[h]
  \centering
  \caption{Pedestrian detection results. $MR^{-2}$ is used to compare the performance of the detectors (the lower the better). "Reasonable", "Heavy", "Partial", and "Bare" are different subsets that are defined by Citypersons validation dataset according to occlusion ratio. The best performances are in \textbf{bold}.}
\begin{tabular}{c|c|c|ccc}
  \hline
  Method& Scale& Reasonable & Heavy & Partial & Bare \\
  \hline
  \multirow{2}{1cm}{\cite{zhang2017citypersons}}& \multirow{2}{0.5cm}{$\times 1$\\$\times 1.3$}& \multirow{2}{0.5cm}{15.4\\ 12.8} & \multirow{2}{0.4cm}{-\\-} & \multirow{2}{0.4cm}{ -\\ \ - } & \multirow{2}{0.4cm}{-\\-} \\
  \multirow{2}{0.5cm}{}& \multirow{2}{0.5cm}{}& \multirow{2}{0.5cm}{} & \multirow{2}{0.5cm}{} & \multirow{2}{0.5cm}{} & \multirow{2}{0.5cm}{}\\
  \hline
  \multirow{2}{1cm}{\cite{zhang2018occlusion}}& \multirow{2}{0.5cm}{$\times 1$\\$\times 1.3$}& \multirow{2}{0.5cm}{12.8\\ 11.0} & \multirow{2}{0.4cm}{55.7\\51.3} & \multirow{2}{0.4cm}{15.3\\13.7 } & \multirow{2}{0.4cm}{6.7\\5.9} \\
  \multirow{2}{0.5cm}{}& \multirow{2}{0.5cm}{}& \multirow{2}{0.5cm}{} & \multirow{2}{0.5cm}{} & \multirow{2}{0.5cm}{} & \multirow{2}{0.5cm}{}\\
  \hline
  \multirow{3}{1cm}{\cite{wang2017repulsion}}& \multirow{3}{0.5cm}{$\times 1$\\$\times 1.3$\\$\times 1.5$}& \multirow{3}{0.5cm}{13.2\\ 11.6\\10.9} & \multirow{3}{0.4cm}{56.9\\55.3\\52.9} & \multirow{2}{0.4cm}{16.8\\14.8\\13.4 } & \multirow{2}{0.4cm}{7.6\\7.0\\6.3} \\
  \multirow{2}{0.5cm}{}& \multirow{2}{0.5cm}{}& \multirow{2}{0.5cm}{} & \multirow{2}{0.5cm}{} & \multirow{2}{0.5cm}{} & \multirow{2}{0.5cm}{}\\
  \multirow{2}{0.5cm}{}& \multirow{2}{0.5cm}{}& \multirow{2}{0.5cm}{} & \multirow{2}{0.5cm}{} & \multirow{2}{0.5cm}{} & \multirow{2}{0.5cm}{}\\
  \hline
  \multirow{3}{1cm}{Ours}& \multirow{3}{0.5cm}{$\times 1$\\$\times 1.3$\\$\times 1.5$}& \multirow{3}{0.5cm}{12.6\\11.4\\10.7} & \multirow{3}{0.4cm}{54.3\\51.1\\49.9} & \multirow{2}{0.4cm}{12.8\\11.3\\10.7 } & \multirow{2}{0.4cm}{7.4\\6.6\\6.5} \\
  \multirow{2}{0.5cm}{}& \multirow{2}{0.5cm}{}& \multirow{2}{0.5cm}{} & \multirow{2}{0.5cm}{} & \multirow{2}{0.5cm}{} & \multirow{2}{0.5cm}{}\\
  \multirow{2}{0.5cm}{}& \multirow{2}{0.5cm}{}& \multirow{2}{0.5cm}{} & \multirow{2}{0.5cm}{} & \multirow{2}{0.5cm}{} & \multirow{2}{0.5cm}{}\\
  \hline
  \multirow{3}{1cm}{Ours + GAN}& \multirow{3}{0.5cm}{$\times 1$\\$\times 1.3$\\$\times 1.5$}& \multirow{3}{0.5cm}{12.4\\11.2\\\textbf{10.5}} & \multirow{3}{0.4cm}{53.8\\50.7\\\textbf{49.3}} & \multirow{2}{0.4cm}{12.7\\11.6\\ \textbf{10.4} } & \multirow{2}{0.4cm}{7.2\\6.5\\ \textbf{6.4}} \\
  \multirow{2}{0.5cm}{}& \multirow{2}{0.5cm}{}& \multirow{2}{0.5cm}{} & \multirow{2}{0.5cm}{} & \multirow{2}{0.5cm}{} & \multirow{2}{0.5cm}{}\\
  \multirow{2}{0.5cm}{}& \multirow{2}{0.5cm}{}& \multirow{2}{0.5cm}{} & \multirow{2}{0.5cm}{} & \multirow{2}{0.5cm}{} & \multirow{2}{0.5cm}{}\\
  \hline
\end{tabular}
\label{tab 3}
\end{table}
\vspace{-0.3cm}

\section{Conclusion}
We propose a multi-modal cascaded generative adversarial networks (PMC-GANs) to synthesize pedestrian images. The model uses multi-scale residual blocks in the encoder part of the generator to obtain multi-scale representation of pedestrian images and uses channel attention residual blocks in the decoder part of the generator to help select the most important features. Our model dramatically outperforms baselines in generating both realistic and diversified pedestrian images, especially in producing high-resolution ones. The experiment of using the PMC-GANs to augment pedestrian detection data further proves its effectiveness and applicability. However, sometimes the direction of light that falls on the generated pedestrian does not match the lightening condition of the background image, which looks artificial. We plan to fix the problem by taking into account lighting variables as future work.\footnote{Please contact wujie@cetc-cloud.com for more information and the supplementary material.}

\bibliography{bmvc_final_png2jpg}

\begin{thebibliography}{48}
\providecommand{\natexlab}[1]{#1}
\providecommand{\url}[1]{\texttt{#1}}
\expandafter\ifx\csname urlstyle\endcsname\relax
  \providecommand{\doi}[1]{doi: #1}\else
  \providecommand{\doi}{doi: \begingroup \urlstyle{rm}\Url}\fi

\bibitem[Almahairi et~al.(2018)Almahairi, Rajeswar, Sordoni, Bachman, and
  Courville]{almahairi2018augmented}
Amjad Almahairi, Sai Rajeswar, Alessandro Sordoni, Philip Bachman, and Aaron
  Courville.
\newblock Augmented cyclegan: Learning many-to-many mappings from unpaired
  data.
\newblock \emph{arXiv preprint arXiv:1802.10151}, 2018.

\bibitem[Brock et~al.(2018)Brock, Donahue, and Simonyan]{brock2018large}
Andrew Brock, Jeff Donahue, and Karen Simonyan.
\newblock Large scale gan training for high fidelity natural image synthesis.
\newblock \emph{arXiv preprint arXiv:1809.11096}, 2018.

\bibitem[Chen et~al.(2016)Chen, Duan, Houthooft, Schulman, Sutskever, and
  Abbeel]{chen2016infogan}
Xi~Chen, Yan Duan, Rein Houthooft, John Schulman, Ilya Sutskever, and Pieter
  Abbeel.
\newblock Infogan: Interpretable representation learning by information
  maximizing generative adversarial nets.
\newblock In \emph{Advances in neural information processing systems}, pages
  2172--2180, 2016.

\bibitem[Choi et~al.(2017)Choi, Choi, Kim, Ha, Kim, and Choo]{choi2017stargan}
Yunjey Choi, Minje Choi, Munyoung Kim, Jung-Woo Ha, Sunghun Kim, and Jaegul
  Choo.
\newblock Stargan: Unified generative adversarial networks for multi-domain
  image-to-image translation.
\newblock \emph{arXiv preprint arXiv:1711.09020}, 2017.

\bibitem[Cordts et~al.(2016)Cordts, Omran, Ramos, Rehfeld, Enzweiler, Benenson,
  Franke, Roth, and Schiele]{Cordts2016Cityscapes}
Marius Cordts, Mohamed Omran, Sebastian Ramos, Timo Rehfeld, Markus Enzweiler,
  Rodrigo Benenson, Uwe Franke, Stefan Roth, and Bernt Schiele.
\newblock The cityscapes dataset for semantic urban scene understanding.
\newblock In \emph{Proc. of the IEEE Conference on Computer Vision and Pattern
  Recognition (CVPR)}, 2016.

\bibitem[Doll{\'a}r et~al.(2009{\natexlab{a}})Doll{\'a}r, Tu, Perona, and
  Belongie]{dollar2009integral}
Piotr Doll{\'a}r, Zhuowen Tu, Pietro Perona, and Serge Belongie.
\newblock Integral channel features.
\newblock 2009{\natexlab{a}}.

\bibitem[Doll{\'a}r et~al.(2009{\natexlab{b}})Doll{\'a}r, Wojek, Schiele, and
  Perona]{dollar2009pedestrian}
Piotr Doll{\'a}r, Christian Wojek, Bernt Schiele, and Pietro Perona.
\newblock Pedestrian detection: A benchmark.
\newblock In \emph{Computer Vision and Pattern Recognition, 2009. CVPR 2009.
  IEEE Conference on}, pages 304--311. IEEE, 2009{\natexlab{b}}.

\bibitem[Doll\'ar et~al.(2012)Doll\'ar, Wojek, Schiele, and
  Perona]{Dollar2012PAMI}
Piotr Doll\'ar, Christian Wojek, Bernt Schiele, and Pietro Perona.
\newblock Pedestrian detection: An evaluation of the state of the art.
\newblock \emph{PAMI}, 34, 2012.

\bibitem[Doll{\'a}r et~al.(2014)Doll{\'a}r, Appel, Belongie, and
  Perona]{dollar2014fast}
Piotr Doll{\'a}r, Ron Appel, Serge Belongie, and Pietro Perona.
\newblock Fast feature pyramids for object detection.
\newblock \emph{IEEE Transactions on Pattern Analysis and Machine
  Intelligence}, 36\penalty0 (8):\penalty0 1532--1545, 2014.

\bibitem[Donahue et~al.(2016)Donahue, Kr{\"a}henb{\"u}hl, and
  Darrell]{donahue2016adversarial}
Jeff Donahue, Philipp Kr{\"a}henb{\"u}hl, and Trevor Darrell.
\newblock Adversarial feature learning.
\newblock \emph{arXiv preprint arXiv:1605.09782}, 2016.

\bibitem[Dowson and Landau(1982)]{dowson1982frechet}
DC~Dowson and BV~Landau.
\newblock The fr{\'e}chet distance between multivariate normal distributions.
\newblock \emph{Journal of multivariate analysis}, 12\penalty0 (3):\penalty0
  450--455, 1982.

\bibitem[Dumoulin et~al.(2016)Dumoulin, Belghazi, Poole, Mastropietro, Lamb,
  Arjovsky, and Courville]{dumoulin2016adversarially}
Vincent Dumoulin, Ishmael Belghazi, Ben Poole, Olivier Mastropietro, Alex Lamb,
  Martin Arjovsky, and Aaron Courville.
\newblock Adversarially learned inference.
\newblock \emph{arXiv preprint arXiv:1606.00704}, 2016.

\bibitem[Frid-Adar et~al.(2018)Frid-Adar, Klang, Amitai, Goldberger, and
  Greenspan]{frid2018synthetic}
Maayan Frid-Adar, Eyal Klang, Michal Amitai, Jacob Goldberger, and Hayit
  Greenspan.
\newblock Synthetic data augmentation using gan for improved liver lesion
  classification.
\newblock In \emph{Biomedical Imaging (ISBI 2018), 2018 IEEE 15th International
  Symposium on}, pages 289--293. IEEE, 2018.

\bibitem[Ge et~al.(2018)Ge, Li, Zhao, Yin, Yi, Wang, and Li]{ge2018fd}
Yixiao Ge, Zhuowan Li, Haiyu Zhao, Guojun Yin, Shuai Yi, Xiaogang Wang, and
  Hongsheng Li.
\newblock Fd-gan: Pose-guided feature distilling gan for robust person
  re-identification.
\newblock \emph{arXiv preprint arXiv:1810.02936}, 2018.

\bibitem[Geiger et~al.(2012)Geiger, Lenz, and Urtasun]{Geiger2012CVPR}
Andreas Geiger, Philip Lenz, and Raquel Urtasun.
\newblock Are we ready for autonomous driving? the kitti vision benchmark
  suite.
\newblock In \emph{Conference on Computer Vision and Pattern Recognition
  (CVPR)}, 2012.

\bibitem[Girshick(2015)]{girshick2015fast}
Ross Girshick.
\newblock Fast r-cnn.
\newblock In \emph{Proceedings of the IEEE international conference on computer
  vision}, pages 1440--1448, 2015.

\bibitem[Goodfellow et~al.(2014)Goodfellow, Pouget-Abadie, Mirza, Xu,
  Warde-Farley, Ozair, Courville, and Bengio]{goodfellow2014generative}
Ian Goodfellow, Jean Pouget-Abadie, Mehdi Mirza, Bing Xu, David Warde-Farley,
  Sherjil Ozair, Aaron Courville, and Yoshua Bengio.
\newblock Generative adversarial nets.
\newblock In \emph{Advances in neural information processing systems}, pages
  2672--2680, 2014.

\bibitem[He et~al.(2014)He, Zhang, Ren, and Sun]{he2014spatial}
Kaiming He, Xiangyu Zhang, Shaoqing Ren, and Jian Sun.
\newblock Spatial pyramid pooling in deep convolutional networks for visual
  recognition.
\newblock In \emph{European conference on computer vision}, pages 346--361.
  Springer, 2014.

\bibitem[He et~al.(2016)He, Zhang, Ren, and Sun]{he2016deep}
Kaiming He, Xiangyu Zhang, Shaoqing Ren, and Jian Sun.
\newblock Deep residual learning for image recognition.
\newblock In \emph{Proceedings of the IEEE conference on computer vision and
  pattern recognition}, pages 770--778, 2016.

\bibitem[Heusel et~al.(2017)Heusel, Ramsauer, Unterthiner, Nessler, and
  Hochreiter]{heusel2017gans}
Martin Heusel, Hubert Ramsauer, Thomas Unterthiner, Bernhard Nessler, and Sepp
  Hochreiter.
\newblock Gans trained by a two time-scale update rule converge to a local nash
  equilibrium.
\newblock In \emph{Advances in Neural Information Processing Systems}, pages
  6626--6637, 2017.

\bibitem[Hosang et~al.(2015)Hosang, Omran, Benenson, and
  Schiele]{hosang2015taking}
Jan Hosang, Mohamed Omran, Rodrigo Benenson, and Bernt Schiele.
\newblock Taking a deeper look at pedestrians.
\newblock In \emph{Proceedings of the IEEE Conference on Computer Vision and
  Pattern Recognition}, pages 4073--4082, 2015.

\bibitem[Isola et~al.(2017)Isola, Zhu, Zhou, and Efros]{isola2017image}
Phillip Isola, Jun-Yan Zhu, Tinghui Zhou, and Alexei~A Efros.
\newblock Image-to-image translation with conditional adversarial networks.
\newblock In \emph{2017 IEEE Conference on Computer Vision and Pattern
  Recognition (CVPR)}, pages 5967--5976. IEEE, 2017.

\bibitem[Kinga and Adam(2015)]{kinga2015method}
D~Kinga and J~Ba Adam.
\newblock A method for stochastic optimization.
\newblock In \emph{International Conference on Learning Representations
  (ICLR)}, volume~5, 2015.

\bibitem[Kingma and Welling(2014)]{kingma2014auto}
Diederik~P Kingma and Max Welling.
\newblock Auto-encoding variational bayes.
\newblock \emph{stat}, 1050:\penalty0 10, 2014.

\bibitem[Larsen et~al.(2016)Larsen, S{\o}nderby, Larochelle, and
  Winther]{larsen2016autoencoding}
Anders Boesen~Lindbo Larsen, S{\o}ren~Kaae S{\o}nderby, Hugo Larochelle, and
  Ole Winther.
\newblock Autoencoding beyond pixels using a learned similarity metric.
\newblock In \emph{International Conference on Machine Learning}, pages
  1558--1566, 2016.

\bibitem[Li et~al.(2018)Li, Fang, Mei, and Zhang]{li2018multi}
Juncheng Li, Faming Fang, Kangfu Mei, and Guixu Zhang.
\newblock Multi-scale residual network for image super-resolution.
\newblock In \emph{Proceedings of the European Conference on Computer Vision
  (ECCV)}, pages 517--532, 2018.

\bibitem[Lin et~al.(2018)Lin, Xia, Qin, Chen, and Liu]{lin2018conditional}
Jianxin Lin, Yingce Xia, Tao Qin, Zhibo Chen, and Tie-Yan Liu.
\newblock Conditional image-to-image translation.
\newblock In \emph{The IEEE Conference on Computer Vision and Pattern
  Recognition (CVPR)(July 2018)}, 2018.

\bibitem[Mao et~al.(2017{\natexlab{a}})Mao, Xiao, Jiang, and Cao]{mao2017can}
Jiayuan Mao, Tete Xiao, Yuning Jiang, and Zhimin Cao.
\newblock What can help pedestrian detection?
\newblock In \emph{2017 IEEE Conference on Computer Vision and Pattern
  Recognition (CVPR)}, pages 6034--6043. IEEE, 2017{\natexlab{a}}.

\bibitem[Mao et~al.(2017{\natexlab{b}})Mao, Li, Xie, Lau, Wang, and
  Smolley]{mao2017least}
Xudong Mao, Qing Li, Haoran Xie, Raymond~YK Lau, Zhen Wang, and Stephen~Paul
  Smolley.
\newblock Least squares generative adversarial networks.
\newblock In \emph{Computer Vision (ICCV), 2017 IEEE International Conference
  on}, pages 2813--2821. IEEE, 2017{\natexlab{b}}.

\bibitem[Mariani et~al.(2018)Mariani, Scheidegger, Istrate, Bekas, and
  Malossi]{mariani2018bagan}
Giovanni Mariani, Florian Scheidegger, Roxana Istrate, Costas Bekas, and
  Cristiano Malossi.
\newblock Bagan: Data augmentation with balancing gan.
\newblock \emph{arXiv preprint arXiv:1803.09655}, 2018.

\bibitem[Mirza and Osindero(2014)]{mirza2014conditional}
Mehdi Mirza and Simon Osindero.
\newblock Conditional generative adversarial nets.
\newblock \emph{arXiv preprint arXiv:1411.1784}, 2014.

\bibitem[Ouyang et~al.(2018)Ouyang, Cheng, Jiang, Li, and Zhou]{PSGAN}
Xi~Ouyang, Yu~Cheng, Yifan Jiang, Chun-Liang Li, and Pan Zhou.
\newblock Pedestrian-synthesis-gan: Generating pedestrian data in real scene
  and beyond.
\newblock \emph{arXiv preprint arXiv:1804.02047}, 2018.

\bibitem[Ronneberger et~al.(2015)Ronneberger, Fischer, and
  Brox]{ronneberger2015u}
Olaf Ronneberger, Philipp Fischer, and Thomas Brox.
\newblock U-net: Convolutional networks for biomedical image segmentation.
\newblock In \emph{International Conference on Medical image computing and
  computer-assisted intervention}, pages 234--241. Springer, 2015.

\bibitem[Russakovsky et~al.(2015)Russakovsky, Deng, Su, Krause, Satheesh, Ma,
  Huang, Karpathy, Khosla, Bernstein, et~al.]{russakovsky2015imagenet}
Olga Russakovsky, Jia Deng, Hao Su, Jonathan Krause, Sanjeev Satheesh, Sean Ma,
  Zhiheng Huang, Andrej Karpathy, Aditya Khosla, Michael Bernstein, et~al.
\newblock Imagenet large scale visual recognition challenge.
\newblock \emph{International Journal of Computer Vision}, 115\penalty0
  (3):\penalty0 211--252, 2015.

\bibitem[Simonyan and Zisserman(2014)]{simonyan2014very}
Karen Simonyan and Andrew Zisserman.
\newblock Very deep convolutional networks for large-scale image recognition.
\newblock \emph{arXiv preprint arXiv:1409.1556}, 2014.

\bibitem[Song et~al.(2019)Song, Zhang, Liu, and Mei]{song2019unsupervised}
Sijie Song, Wei Zhang, Jiaying Liu, and Tao Mei.
\newblock Unsupervised person image generation with semantic parsing
  transformation.
\newblock \emph{arXiv preprint arXiv:1904.03379}, 2019.

\bibitem[Tripathi et~al.(2019)Tripathi, Chandra, Agrawal, Tyagi, Rehg, and
  Chari]{tripathi2019learning}
Shashank Tripathi, Siddhartha Chandra, Amit Agrawal, Ambrish Tyagi, James~M
  Rehg, and Visesh Chari.
\newblock Learning to generate synthetic data via compositing.
\newblock \emph{arXiv preprint arXiv:1904.05475}, 2019.

\bibitem[Wang et~al.(2017)Wang, Liu, Zhu, Tao, Kautz, and
  Catanzaro]{wang2017high}
Ting-Chun Wang, Ming-Yu Liu, Jun-Yan Zhu, Andrew Tao, Jan Kautz, and Bryan
  Catanzaro.
\newblock High-resolution image synthesis and semantic manipulation with
  conditional gans.
\newblock \emph{arXiv preprint arXiv:1711.11585}, 2017.

\bibitem[Wang et~al.()Wang, Xiao, Jiang, Shao, Sun, and
  Shen]{wang2017repulsion}
Xinlong Wang, Tete Xiao, Yuning Jiang, Shuai Shao, Jian Sun, and Chunhua Shen.
\newblock Repulsion loss: Detecting pedestrians in a crowd.

\bibitem[Zhang et~al.(2018{\natexlab{a}})Zhang, Goodfellow, Metaxas, and
  Odena]{zhang2018self}
Han Zhang, Ian Goodfellow, Dimitris Metaxas, and Augustus Odena.
\newblock Self-attention generative adversarial networks.
\newblock \emph{arXiv preprint arXiv:1805.08318}, 2018{\natexlab{a}}.

\bibitem[Zhang et~al.(2016)Zhang, Benenson, Omran, Hosang, and
  Schiele]{zhang2016far}
Shanshan Zhang, Rodrigo Benenson, Mohamed Omran, Jan Hosang, and Bernt Schiele.
\newblock How far are we from solving pedestrian detection?
\newblock In \emph{Proceedings of the IEEE Conference on Computer Vision and
  Pattern Recognition}, pages 1259--1267, 2016.

\bibitem[Zhang et~al.(2017)Zhang, Benenson, and Schiele]{zhang2017citypersons}
Shanshan Zhang, Rodrigo Benenson, and Bernt Schiele.
\newblock Citypersons: A diverse dataset for pedestrian detection.
\newblock In \emph{The IEEE Conference on Computer Vision and Pattern
  Recognition (CVPR)}, volume~1, page~3, 2017.

\bibitem[Zhang et~al.(2018{\natexlab{b}})Zhang, Wen, Bian, Lei, and
  Li]{zhang2018occlusion}
Shifeng Zhang, Longyin Wen, Xiao Bian, Zhen Lei, and Stan~Z Li.
\newblock Occlusion-aware r-cnn: Detecting pedestrians in a crowd.
\newblock In \emph{European Conference on Computer Vision}, pages 657--674.
  Springer, 2018{\natexlab{b}}.

\bibitem[Zhang et~al.(2018{\natexlab{c}})Zhang, Li, Li, Wang, Zhong, and
  Fu]{zhang2018image}
Yulun Zhang, Kunpeng Li, Kai Li, Lichen Wang, Bineng Zhong, and Yun Fu.
\newblock Image super-resolution using very deep residual channel attention
  networks.
\newblock \emph{arXiv preprint arXiv:1807.02758}, 2018{\natexlab{c}}.

\bibitem[Zhu et~al.(2017{\natexlab{a}})Zhu, Park, Isola, and
  Efros]{zhu2017unpaired}
Jun-Yan Zhu, Taesung Park, Phillip Isola, and Alexei~A Efros.
\newblock Unpaired image-to-image translation using cycle-consistent
  adversarial networks.
\newblock In \emph{Computer Vision (ICCV), 2017 IEEE International Conference
  on}, pages 2242--2251. IEEE, 2017{\natexlab{a}}.

\bibitem[Zhu et~al.(2017{\natexlab{b}})Zhu, Zhang, Pathak, Darrell, Efros,
  Wang, and Shechtman]{zhu2017toward}
Jun-Yan Zhu, Richard Zhang, Deepak Pathak, Trevor Darrell, Alexei~A Efros,
  Oliver Wang, and Eli Shechtman.
\newblock Toward multimodal image-to-image translation.
\newblock In \emph{Advances in Neural Information Processing Systems}, pages
  465--476, 2017{\natexlab{b}}.

\bibitem[Zhu et~al.(2018)Zhu, Aoun, Krijn, and Vanschoren]{ZhuAKV18}
Yezi Zhu, Marc Aoun, Marcel Krijn, and Joaquin Vanschoren.
\newblock Data augmentation using conditional generative adversarial networks
  for leaf counting in arabidopsis plants.
\newblock In \emph{British Machine Vision Conference 2018, {BMVC} 2018,
  Northumbria University, Newcastle, UK, September 3-6, 2018}, page 324, 2018.
\newblock URL \url{http://bmvc2018.org/contents/workshops/cvppp2018/0014.pdf}.

\bibitem[Zhu et~al.(2019)Zhu, Huang, Shi, Yu, Wang, and
  Bai]{zhu2019progressive}
Zhen Zhu, Tengteng Huang, Baoguang Shi, Miao Yu, Bofei Wang, and Xiang Bai.
\newblock Progressive pose attention transfer for person image generation.
\newblock \emph{arXiv preprint arXiv:1904.03349}, 2019.

\end{thebibliography}

\end{document}